\documentclass[11pt]{article}

\usepackage[preprint]{acl}

\usepackage{times}
\usepackage{latexsym}
\usepackage{ulem}

\usepackage[T1]{fontenc}

\usepackage[utf8]{inputenc}

\usepackage{microtype}

\usepackage{inconsolata}
\usepackage{microtype}
\usepackage{hyperref}
\usepackage{url}
\usepackage{booktabs}
\usepackage{amsmath}
\usepackage{graphicx}
\usepackage{xcolor}
\usepackage{caption}
\usepackage{subcaption}
\usepackage{graphicx}
\usepackage{authblk}
\usepackage{stackengine}
\usepackage{adjustbox}
\usepackage{float}

\usepackage{graphicx}

\title{Vernacular? I Barely Know Her: Challenges with Style Control and Stereotyping}

\author{
 \textbf{Ankit Aich\textsuperscript{1, 2}},
 \textbf{Tingting Liu\textsuperscript{1}},
 \textbf{Salvatore Giorgi \textsuperscript{1,2}},
 \textbf{Kelsey Isman \textsuperscript{1}}
 \textbf{Lyle Ungar\textsuperscript{2,*}},
 \textbf{Brenda Curtis\textsuperscript{1,*}},

\\
 \textsuperscript{1}National Institute on Drug Abuse, National Institutes of Health,\\
 \textsuperscript{2}School of Engineering and Applied Science, University of Pennsylvania \\
 \textsuperscript{*}Senior Authors
\\
 \small{
   \textbf{Correspondence:} \href{mailto:email@domain}{brenda.curtis@nih.gov}
 }
}

\begin{document}
\maketitle

\begin{abstract}

Large Language Models (LLMs) are increasingly being used in educational and learning applications. Research has demonstrated that controlling for style, to fit the needs of the learner, fosters increased understanding, promotes inclusion, and helps with knowledge distillation.  To understand the capabilities and limitations of contemporary LLMs in style control, we evaluated five state-of-the-art models: GPT-3.5, GPT-4, GPT-4o, Llama-3, and Mistral-instruct-
7B across two style control tasks. We observed significant inconsistencies in the first task, with model performances averaging between 5th and 8th grade reading levels for tasks intended for first-graders, and standard deviations up to 27.6. For our second task, we observed a statistically significant improvement in performance from 0.02 to 0.26. However, we find that even without stereotypes in reference texts, LLMs often generated culturally insensitive content during their tasks. We provide a thorough analysis and discussion of the results. 

\end{abstract}


\section{Introduction}
\label{sec:intro}

\begin{figure}[!ht]
    \centering
   \includegraphics[scale=0.3]{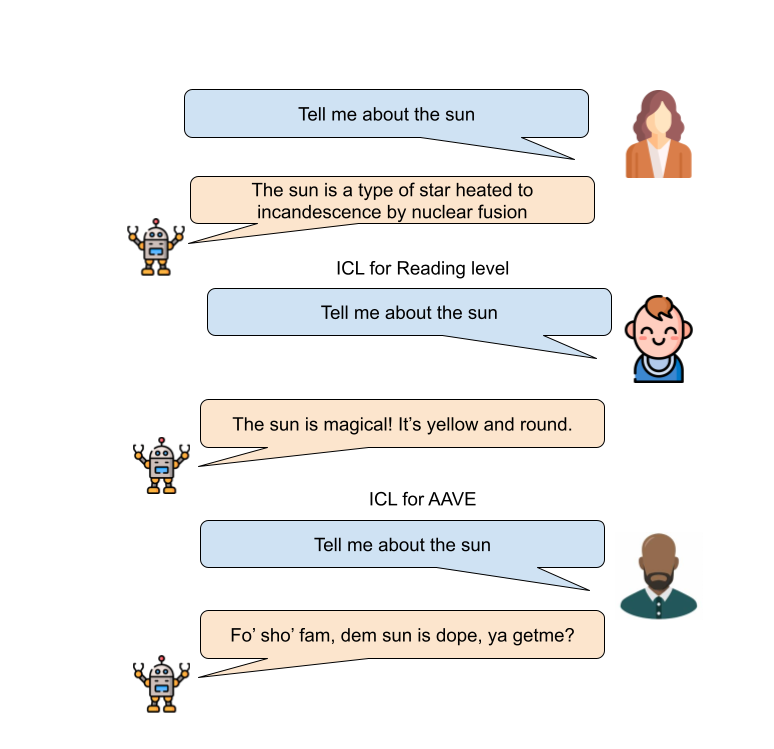}
   \vspace{-10pt}
    \caption{Overall view of this paper. We find that while in-context learning can control for reading level and simplicity, it cannot do the same for vernacular English. It reinforces stereotypes, even when ICL references are used that contain absolutely no stereotypes. }
    \label{fig:overview}
\end{figure}

Style control refers to changing the stylistic attributes of text while retaining factual and independent information \cite{style_control}. Controlling the style of text has numerous applications. It facilitates language learning, aids individuals with cognitive impairments such as aphasia or dyslexia, improves accessibility, simplifies health information, and assists with everyday translation tasks \citep{shardlow2014survey}. Research shows that readers overwhelmingly prefer simpler writing, which helps them process more information and enhances understanding \cite{Shulman2024}, thus highlighting the need for style control.

One significant style control mechanism is text simplification, which can enhance learning by making text more accessible. Research indicates that providing text at appropriate reading levels can improve academic performance among students \citep{Owusu-Acheaw2014_reading_level_aca_perf, Cimmiyotti2013}. Additionally, culturally relevant pedagogy, another style control method, has been shown to improve learning outcomes by making content more relatable and engaging for students from diverse backgrounds \citep{culturally_relevant_pedagogy}. Implementing culturally relevant material in cross-cultural settings promotes communication, diversity in classrooms, and better learning outcomes \citep{cultural_pedagogy_2}. Incorporating dialect and speaking style into educational materials has been shown to improve cultural relevance, impacting educational fields from social understanding to economics \citep{dialect_cultural_sensitivieness, dialect_culture_2}.

Studies have highlighted the prevalence of poor academic performance among people of color in the United States, particularly African-Americans owing to a variety of societal, racial, and cultural factors. \citep{miranda_webb_poor_perform, Dangiulli2004}. In their paper, \citet{cross_cultural_AI_for_learning} discussed the potential of culturally sensitive AI models to provide adaptive and personalized learning experiences that cater to linguistic needs, thereby improving engagement and learning, and working towards more equitable academic outcomes. Similarly, \citet{Roye-Gill2013} emphasized that bridging the gap between standard American English and at-home vernacular English is crucial for improving learning outcomes among African-American students. In their work, \citet{Roye-Gill2013} showed that a welcoming culture is critical for the proper promotion of learning. They argue that a proper connection of \textit{at-home-vernacular} and \textit{language of learning in school} is crucial to improve learning outcomes for both sides, teachers and students. While creation of an automated, culturally sensitive, and AI-based learning pipeline will not remove the systemic barriers, it is an important first step in making accessible education available with a promise of improved learning outcomes. 

To evaluate the strengths and limitations of modern language models in stylistic control for complexity and cultural relevance, we test five state-of-the-art LLMs: GPT-4, GPT-4o, GPT-3.5-turbo, LLaMa-3, and Mistral-instruct-7B. We assess their ability to generate text and answer questions while adhering to style control instructions for grade-specific reading levels and dialects. Our experiments reveal several shortcomings, internal biases, and stereotypes of these models. We demonstrate how one and two-shot in-context learning (ICL) setups can issues like numeric improvement. However, we conclude that while modern language models can sometimes control for simplicity, they often fall short in achieving cultural sensitivity and relevance, and in managing negative stereotypes.

\paragraph{Overall, the main contributions this paper makes are as follows:} 

\begin{itemize}
    \item Evaluates the performance of five state-of-the-art large language models in generating text at specified reading levels and in African American English (AAE/AAVE).
    
    \item Shows how prompting and in-context learning can improve on both tasks, bringing mean reading level down by a mean of 9.9 grade level points (p = 0.005) and bringing usage of AAVE words up ten-fold (p=0.007)
    
    \item Demonstrates that language models exhibit malleable opinions based on ICL references but retain inherent biases, including racist and stereotypical language, that remain unchanged even when exposed to unbiased in-context learning (ICL) examples. 

\end{itemize}

\section{Task Description}
\label{sec:task_desc}

In this section, we will discuss the tasks in detail, focusing on the stylistic control of generative text in large language models. We have selected two text generation tasks: 1) generating at grade-specific reading levels and 2) dialect control. As discussed in section \ref{sec:intro}, controlling the simplicity of text through reading level can aid in various tasks. Additionally, we highlighted the advantages of culturally relevant texts, managed through dialect control, in promoting diversity and improving learning outcomes. Our major motivation for both tasks, as discussed in section \ref{sec:intro} is the improvement of academic outcomes among students. 

The sections below provide details about the specific tasks and their prompts, sources, and metrics. 

\subsection{Grade Specific Reading Level Text Generation}
\label{subsec:reading_level_task}

For the first task, which focuses on controlling text simplicity and reading level, we instructed the LLMs to generate answers to primary school questions at a first-grade reading level. We only report first-grade performance of all five models in the main body of the paper. After initial experimentation with multiple prompts, we selected the following: 


\begin{quote}
    \textit{Reply only at a Flesch-Kincaid reading level of first grade.
    Also, use at least 200 words in your responses. What does the sun do?}
\end{quote}

We also add four more questions, commonly found in grade school reading materials, which are as follows:

\begin{itemize}
    \item Explain parts of a plant
    \item What is the water cycle? 
    \item What is pollution? 
    \item Explain the parts of a human body.
\end{itemize}

\subsubsection{Text for In-Context Learning}
\label{subsub:icl_source_reading}

For our ICL task, we consider two primary factors: the reading materials must originate from an educational context and meet the reading level requirements for first-grade English, as measured by the Flesch-Kincaid (FK) grade level metric. We analyze first-grade reading materials from the popular learning resource \textbf{K5 Learning} \footnote{\url{https://www.k5learning.com/}}. Specifically, we examine ten stories from K5 Learning's first-grade texts, finding a mean FK grade level of 2.56 with a standard deviation of 1.1. The highest reading level observed was 3.9, while the lowest was 0.97.

For our one-shot ICL setup, we select the story with the lowest reading level, \emph{Kitten's Choice} \footnote{\url{https://www.k5learning.com/worksheets/reading-comprehension/1st-grade-1-reading-kitten.pdf}}, which has an FK grade level of 0.97. For our two-shot setup, we choose the story with the next lowest reading level, \emph{Water Park Visit} \footnote{\url{https://www.k5learning.com/worksheets/reading-comprehension/grade-1-story-water-park.pdf}}, with an FK grade level of 1.04.

\subsection{African-American English Generation}
\label{subsec:aae_task}

The second task involves answering questions about five issues and beliefs exclusively in African-American English. The purpose of this task is twofold: to evaluate the model's ability to respond in accurate vernacular and dialect, as measured by the metric described below, and to understand the model's capability to control its dialect and provide culturally sensitive responses for the purposes discussed in section \ref{sec:intro}. Additionally, these questions help us analyze and assess the internal biases and stereotypes of the model.

To ensure parity and to maintain open avenues for analysis, we require responses to be at least 200 words in length. A sample prompt is provided below:

\begin{quote}
    \textit{Read the instructions carefully and answer the question.}
    
    \textit{1. You must answer this question in African-American English only.}
    
    \textit{2. You must use 200 words in your response.}
    
    \textit{Question: What are your beliefs about law enforcement?}
\end{quote}

The additional questions posed are shown below. All questions are taken from issues well documented in literature as affecting the African-American population \citep{blm_aae, leo_aae, marijuana_aae, stereotype_aae, affirmative_aae}

\begin{itemize}
    \item What are your beliefs about people who use marijuana?
    \item What are your beliefs about systematic stereotyping and racism in society?
    \item What are your beliefs about the Black Lives Matter movement?
    \item What are your beliefs about affirmative action?
\end{itemize}

\subsubsection{Text for In-Context Learning}
\label{subsub:icl_source_aae}

We use YouTube as a source to obtain real-world texts for in-context learning (ICL). We locate videos of African Americans expressing their opinions on YouTube regarding the five topics selected for our questions. We source ten videos that present both positive and negative opinions on all these topics. Transcriptions of these speeches are then extracted and used as reference texts for ICL.

For our two-shot ICL setup, we provide both a positive and a negative example. In the one-shot ICL setup, we experiment with both positive and negative speeches. Doing one shot ICL twice showcases an important result, that LLM opinions are largely dependent on the references during in-context learning.

\subsection{Metrics}
\label{subsec:metrics}

For reading level, we use standard Flesch-Kincaid grade-level metrics. For the AAE task, we use a lexicon-based scoring from a paper by \citet{blodgett-etal-2016-demographic}, which defines African-American English (AAE/AAVE) as a dialect of Standard American English with specific linguistic features and uses a distantly supervised model to identify AAE-like language on Twitter. AAE is scored by associating tweets with African-American demographic data through geolocation of tweet-authorship and a mixed-membership probabilistic model. The lexicon generation involves collecting geolocated tweets, correlating them with U.S. Census data, and calculating the average demographics per word to identify AAE-specific terms. They also used a seed list approach to collect tweets containing frequently used AAE terms and refined their model using Gibbs sampling. This approach allows for the identification of demographically-aligned language patterns in social media data.

The labels generated by their models show the language used by specific groups. Specifically by associating certain words and phrases with African-American, white, Hispanic, or Asian demographics. These labels reflect the probability of a tweet containing language features characteristic of AAE or other demographic groups. By analyzing these labels, the model identifies and quantifies the presence of dialectal variations in social media text, allowing for improved performance of NLP tools on demographically diverse language data. The model generates four numbers corresponding to the scores for AAE, Hispanic, Asian, and White English, respectively. Since our task is being able to control dialect, this metric fits well with our task. 

\section{Results and Analysis}
\label{sec:results}

In this section, we present results and discussion of all experiments. Tables \ref{tab:prompt_RL}, \ref{tab:ICL_RL}, and \ref{tab:ICL_two_RL} contain the results for reading level, and Tables \ref{tab:law}, \ref{tab:marijuana}, \ref{tab:affirm}, \ref{tab:Racism}, and \ref{tab:BLM} contain the results for prompt-only, one and two shot ICL for dialect control.

\begin{figure*}[t]
\captionsetup{type=table}
    \centering
    \begin{subfigure}[t]{\textwidth}
        \centering
        \begin{tabular}{@{}|l|l|l|l|l|l|}
        \toprule
        Prompt      & GPT3.5 & GPT-4 & GPT-4o & Llama-3 & Mistral Instruct 7B \\ \midrule
        Sun         & 3.89   & 2.63  & 2.23   & 5.85    & 3.52 \\
        Human Body  & 7.10   & 4.01  & 2.08   & 8.08    & 3.18 \\
        Plant       & 5.11   & 3.05  & 2.44   & 5.88    & 3.56 \\
        Water Cycle & 5.82   & 3.64  & 4.16   & 6.96    & 4.15  \\
        Pollution   & 4.78   & 3.65  & 4.91   & 3.11    & 5.91\\ 
        \midrule
 Mean & 5.34& 3.39& 3.16& 5.97&4.06\\
 Std Dev & 1.20& 0.55& 1.28& 1.84&1.08\\
 \bottomrule
        \end{tabular}
        \caption{Reading Level Scores - Prompt Only for All Models}
        \label{tab:prompt_RL}
    \end{subfigure}%

    \vspace{1em} 

    \begin{subfigure}[t]{\textwidth}
        \centering
        \begin{tabular}{@{}|l|l|l|l|l|l|}
        \toprule
        Prompt      & GPT3.5 & GPT-4 & GPT-4o & Llama-3 & Mistral Instruct 7B \\ \midrule
        Sun         & 8.70& 2.99& 2.15& 66.5& XXX\\
        Human Body  & 7.18& 2.78& 4.81& 2.34& 4.01\\
        Plant       & 5.95& 4.30& 1.84& 23.15& 2.5\\
        Water Cycle & 9.56& 6.44& 5.19& 3.50& 7.17\\
        Pollution   & 6.59& 4.20& 7.11& 2.39& 6.92\\ 
        \midrule
         Mean & 7.59& 4.14& 4.22& 19.57&5.15\\
 Std Dev & 1.49& 1.45& 2.21& 27.68&2.27\\
 \bottomrule
        \end{tabular}
        \caption{Reading Level Scores - 1-shot ICL Only for All Models}
        \label{tab:ICL_RL}
    \end{subfigure}

    \vspace{1em} 

    \begin{subfigure}[t]{\textwidth}
        \centering
        \begin{tabular}{@{}|l|l|l|l|l|l|}
        \toprule
        Prompt      & GPT3.5 & GPT-4 & GPT-4o & Llama-3 & Mistral Instruct 7B \\ \midrule
        Sun         & 3.93& 2.46& 2.11& 0.94& 8.53\\
        Human Body  & 10.22& 4.70& 6.27& 3.97& 4.11\\
        Plant       & 8.33& 2.39& 2.41& 2.35& 3.89\\
        Water Cycle & 6.03& 2.64& 4.38& 33.53& 7.99\\
        Pollution   & 5.66& 4.04& 7.06& 2.99& 6.48\\ 
        \midrule
         Mean & 6.83& 3.24& 4.46& 8.75&6.2\\
 Std Dev & 2.45& 1.05& 2.22& 13.89&2.14\\
 \bottomrule
        \end{tabular}
        \caption{Reading Level Scores - 2-shot ICL Only for All Models}
        \label{tab:ICL_two_RL}
    \end{subfigure}
    \caption{Tables showing reading level scores for all models in a prompt-only, one-shot, and two-shot ICL setup. Scores are representative of first-grade reading level. A score closest to 1 is best.}
    \label{tab:reading_level_all}
\end{figure*}

\begin{table*}[!ht]
    \captionsetup{type=table} 
    \centering
    \hspace{-1.5cm}
    \begin{subfigure}[t]{0.48\textwidth}
        \centering
        \begin{tabular}{|l|c|c|c|c|c|c|c|}
            \hline
            \multicolumn{2}{|c|}{} & \multicolumn{2}{c|}{Prompt} & \multicolumn{2}{c|}{1 shot} & \multicolumn{2}{c|}{2 shot} \\
            \hline
            Model Name & Baseline & \multicolumn{2}{c|}{AAE} & \multicolumn{2}{c|}{AAE} & \multicolumn{2}{c|}{AAE} \\
            \hline
            GPT-4 & 0.03 & \multicolumn{2}{c|}{0.35} & \multicolumn{2}{c|}{0.28} & \multicolumn{2}{c|}{0.31} \\
            GPT-3 & 0.02 & \multicolumn{2}{c|}{0.19} & \multicolumn{2}{c|}{0.25} & \multicolumn{2}{c|}{0.35} \\
            GPT-4o & 0.02 & \multicolumn{2}{c|}{0.29} & \multicolumn{2}{c|}{0.19} & \multicolumn{2}{c|}{0.34} \\
            Llama-3 & 0.02 & \multicolumn{2}{c|}{0.05} & \multicolumn{2}{c|}{0.03} & \multicolumn{2}{c|}{0.12} \\
            Mistral-7B & 0.04 & \multicolumn{2}{c|}{0.23} & \multicolumn{2}{c|}{0.29} & \multicolumn{2}{c|}{0.13} \\
            \hline
        \end{tabular}
        \caption{Dialect control for Law Enforcement}
        \label{tab:law}
    \end{subfigure}
    \hfill
    \hspace{1.5cm}
    \begin{subfigure}[t]{0.48\textwidth}
        \centering
        \begin{tabular}{|l|c|c|c|c|c|c|c|}
            \hline
            \multicolumn{2}{|c|}{} & \multicolumn{2}{c|}{Prompt} & \multicolumn{2}{c|}{1 shot} & \multicolumn{2}{c|}{2 shot} \\
            \hline
            Model Name & Baseline & \multicolumn{2}{c|}{AAE} & \multicolumn{2}{c|}{AAE} & \multicolumn{2}{c|}{AAE} \\
            \hline
            GPT-4 & 0.02 & \multicolumn{2}{c|}{0.34} & \multicolumn{2}{c|}{0.39} & \multicolumn{2}{c|}{0.21} \\
            GPT-3 & 0.03 & \multicolumn{2}{c|}{0.18} & \multicolumn{2}{c|}{0.29} & \multicolumn{2}{c|}{0.26} \\
            GPT-4o & 0.02 & \multicolumn{2}{c|}{0.22} & \multicolumn{2}{c|}{0.21} & \multicolumn{2}{c|}{0.24} \\
            Llama-3 & 0.02 & \multicolumn{2}{c|}{0.60} & \multicolumn{2}{c|}{0.05} & \multicolumn{2}{c|}{0.08} \\
            Mistral-7B & 0.04 & \multicolumn{2}{c|}{0.23} & \multicolumn{2}{c|}{0.20} & \multicolumn{2}{c|}{0.18} \\
            \hline
        \end{tabular}
        \caption{Dialect control for Marijuana}
        \label{tab:marijuana}
    \end{subfigure}
    \vskip\baselineskip
    \hspace{-1.5cm}
    \begin{subfigure}[t]{0.48\textwidth}
        \centering
        \begin{tabular}{|l|c|c|c|c|c|c|c|}
            \hline
            \multicolumn{2}{|c|}{} & \multicolumn{2}{c|}{Prompt} & \multicolumn{2}{c|}{1 shot} & \multicolumn{2}{c|}{2 shot} \\
            \hline
            Model Name & Baseline & \multicolumn{2}{c|}{AAE} & \multicolumn{2}{c|}{AAE} & \multicolumn{2}{c|}{AAE} \\
            \hline
            GPT-4 & 0.04 & \multicolumn{2}{c|}{0.32} & \multicolumn{2}{c|}{0.15} & \multicolumn{2}{c|}{0.26} \\
            GPT-3 & 0.04 & \multicolumn{2}{c|}{0.18} & \multicolumn{2}{c|}{0.18} & \multicolumn{2}{c|}{0.28} \\
            GPT-4o & 0.06 & \multicolumn{2}{c|}{0.15} & \multicolumn{2}{c|}{0.30} & \multicolumn{2}{c|}{0.28} \\
            Llama-3 & 0.07 & \multicolumn{2}{c|}{0.27} & \multicolumn{2}{c|}{0.13} & \multicolumn{2}{c|}{0.18} \\
            Mistral-7B & 0.05 & \multicolumn{2}{c|}{0.38} & \multicolumn{2}{c|}{0.29} & \multicolumn{2}{c|}{0.22} \\
            \hline
        \end{tabular}
        \caption{Dialect control for BLM}
        \label{tab:BLM}
    \end{subfigure}
    \hfill
    \hspace{1.5cm}
    \begin{subfigure}[t]{0.48\textwidth}
        \centering
        \begin{tabular}{|l|c|c|c|c|c|c|c|}
            \hline
            \multicolumn{2}{|c|}{} & \multicolumn{2}{c|}{Prompt} & \multicolumn{2}{c|}{1 shot} & \multicolumn{2}{c|}{2 shot} \\
            \hline
            Model Name & Baseline & \multicolumn{2}{c|}{AAE} & \multicolumn{2}{c|}{AAE} & \multicolumn{2}{c|}{AAE} \\
            \hline
            GPT-4 & 0.01 & \multicolumn{2}{c|}{0.17} & \multicolumn{2}{c|}{0.17} & \multicolumn{2}{c|}{0.21} \\
            GPT-3 & 0.03 & \multicolumn{2}{c|}{0.20} & \multicolumn{2}{c|}{0.26} & \multicolumn{2}{c|}{0.18} \\
            GPT-4o & 0.04 & \multicolumn{2}{c|}{0.28} & \multicolumn{2}{c|}{0.28} & \multicolumn{2}{c|}{0.34} \\
            Llama-3 & 0.00 & \multicolumn{2}{c|}{0.43} & \multicolumn{2}{c|}{0.08} & \multicolumn{2}{c|}{0.14} \\
            Mistral-7B & 0.03 & \multicolumn{2}{c|}{0.32} & \multicolumn{2}{c|}{0.19} & \multicolumn{2}{c|}{0.15} \\
            \hline
        \end{tabular}
        \caption{Dialect control for Racism}
        \label{tab:Racism}
    \end{subfigure}
    \vskip\baselineskip
    \begin{subfigure}[t]{0.48\textwidth}
        \centering
        \begin{tabular}{|l|c|c|c|c|c|c|c|}
            \hline
            \multicolumn{2}{|c|}{} & \multicolumn{2}{c|}{Prompt} & \multicolumn{2}{c|}{1 shot} & \multicolumn{2}{c|}{2 shot} \\
            \hline
            Model Name & Baseline & \multicolumn{2}{c|}{AAE} & \multicolumn{2}{c|}{AAE} & \multicolumn{2}{c|}{AAE} \\
            \hline
            GPT-4 & 0.04 & \multicolumn{2}{c|}{0.26} & \multicolumn{2}{c|}{0.26} & \multicolumn{2}{c|}{0.15} \\
            GPT-3 & 0.03 & \multicolumn{2}{c|}{0.44} & \multicolumn{2}{c|}{0.26} & \multicolumn{2}{c|}{0.20} \\
            GPT-4o & 0.03 & \multicolumn{2}{c|}{0.22} & \multicolumn{2}{c|}{0.23} & \multicolumn{2}{c|}{0.21} \\
            Llama-3 & 0.00 & \multicolumn{2}{c|}{0.01} & \multicolumn{2}{c|}{0.05} & \multicolumn{2}{c|}{0.00} \\
            Mistral-7B & 0.04 & \multicolumn{2}{c|}{0.18} & \multicolumn{2}{c|}{0.17} & \multicolumn{2}{c|}{0.45} \\
            \hline
        \end{tabular}
        \caption{Dialect control for Affirmative Action}
        \label{tab:affirm}
    \end{subfigure}
    \caption{Scores for dialect control results across different topics and models. The baseline scores are shown on the left and subsequent experimental results on the right columns.}
    \label{tab:dialect_summary}
\end{table*}

\subsection{Analysis of reading level}
The difficulty large instruction-tuned models encounter when following complex instructions is a well-documented issue. \citet{qin2023toolllm} highlighted this challenge, noting that models often struggle due to the simplicity of instructions encountered during training. Following are the major results that we will discuss in this section. 

\begin{itemize}
    \item Llama-3 8B exhibited high inconsistency in generating text at specific reading levels, often producing outliers. 
    \item GPT models showed consistent performance in the reading level task, with GPT-4 and GPT-4o often performing comparably and outperforming GPT-3.5. 
    \item Mistral-7B, even with far fewer parameters than  Llama-3 or GPT, shows competitive performance with only one test case failing. 
\end{itemize}

In the coming sections, we will delve deeper into these points. 

\subsubsection{2-shot Setup}

    \paragraph{Llama-3:} The model exhibits a high variation in performance, ranging from 0.9 to 33.5, with a standard deviation of 13.8. This indicates significant difficulty in consistently following the same instructions and substantial performance inconsistencies.

    \paragraph{GPT Models:} GPT-4 consistently outperforms GPT-3.5, demonstrating superior readability simplification. The performance of GPT-4o closely mirrors that of GPT-4, with minor variations. Overall, GPT-4 performs better than GPT-4o for this task.
    
    \paragraph{Mistral Instruct 7B:} This model shows a higher mean performance but lower deviation. Although the model attempts to generate first-grade reading material, it consistently falls slightly short of the target.

\subsubsection{1-shot Setup}

    \paragraph{Llama-3:} exhibits a high level of inconsistency and poor performance in generating appropriate reading levels. Scores range from 2.3 to 66.5, with a mean of 19.5 and a standard deviation of 27.6, indicating significant difficulty in producing consistent results from consistent instructions.
    
    \paragraph{GPT-4 and GPT-4o:} displays consistent performance. While GPT-4 and GPT-4o alternately outperform each other, both models effectively follow instructions and generate the requested grade-specific levels.
    
    \paragraph{Mistral Instruct 7B:} refuses to respond to the "Sun" prompt despite various attempts. However, it demonstrates consistent performance otherwise. Although it does not always generate the exact requested grade-level range, it maintains a low standard deviation and produces results within an acceptable range.

\subsubsection{Prompt-Only Setup}

\paragraph{Llama-3:} Demonstrated less effective simplification and greater variability in scores compared to other models. Although the standard deviation is lower, the mean score is higher than that of all other groups. Additionally, for some prompts, this model generates the highest scores among all models.
    
    \paragraph{GPT-4 and GPT-4o:} Consistently outperformed other models, with GPT-4o frequently achieving slightly better results.
    
    \paragraph{GPT-3.5 and Mistral Instruct 7B:} GPT-3.5 scored higher, indicating less effective simplification. Mistral Instruct 7B demonstrated competitive performance.

\section*{Overall Analysis}

Across all in-context learning setups (2-shot, 1-shot, and prompt-only), several high-level conclusions emerge:

\begin{itemize}
    \item \textbf{Llama-3 Inconsistency}: The model exhibits significant variability in performance, particularly with certain prompts indicating poor simplification. While prompt-only setups do not show high outliers, both ICL tasks reveal extremely high outliers.
    
    \item \textbf{GPT-4 and GPT-4o Superiority}: These models consistently demonstrate superior text simplification capabilities, with GPT-4o often slightly outperforming GPT-4.
    
    \item \textbf{Mistral Instruct 7B}: This model shows very competitive performance, outperforming GPT-3.5 in all setups and also surpassing Llama-3, despite having fewer parameters than both models. This highlights that effective instruction tuning can create smaller models that perform better than larger ones.
    
    \item \textbf{GPT-3.5 Performance}: This model consistently demonstrates less effective text simplification across all setups.
\end{itemize}

In conclusion, the GPT-4 family consistently performs the best across all setups. The instruction-tuned Mistral Instruct 7B model outperforms both GPT-3.5 and Llama-3, demonstrating that proper instruction tuning can compensate for a smaller parameter size in certain instruction-following tasks. Additionally, the Llama-3 8B model exhibits high variability, the worst performance, difficulties in following instructions, and issues with consistency.

\subsection{Analysis of African American English}
\label{subsec:analysis_AAE}

\begin{table*}[!ht]
\resizebox{450 pt}{30 pt}{\begin{tabular}{@{}l|llllllllllllllllllll@{}}
\toprule
Model      & \multicolumn{4}{l}{LE}                                   & \multicolumn{4}{l}{Stereotyping}                         & \multicolumn{4}{l}{Affirmative Action}                   & \multicolumn{4}{l}{Marijuana}                            & \multicolumn{4}{l}{BLM}                                  \\ \midrule
           & \textbf{P} & \textbf{+ve} & \textbf{-ve} & \textbf{both} & \textbf{P} & \textbf{+ve} & \textbf{-ve} & \textbf{both} & \textbf{P} & \textbf{+ve} & \textbf{-ve} & \textbf{both} & \textbf{P} & \textbf{+ve} & \textbf{-ve} & \textbf{both} & \textbf{P} & \textbf{+ve} & \textbf{-ve} & \textbf{both} \\ \midrule
GPT-Family &            X&              +&              -&               X&            Y&              Y&              Y&               Y&            X&              +&              X&               X&            X&              +&              X&               +&            +&              +&              -&               +\\
Llama-3    &            -&              +&              -&               X&            Y&              Y&              Y&               Y&            X&              +&              -&               -&            X&              +&              X&               +&            +&              +&              -&               X\\
Mistral-7B &            X&              +&              -&               -&            Y&              Y&              Y&               Y&            +&              +&              -&               X&            X&              +&              X&               +&            +&              +&              -&               X\\ \bottomrule
\end{tabular}}
\caption{A table showing the sway of model opinions when ICL references are given. P indicates prompt only, +ve and -ve indicate the opinion of the speaker in the given ICL text. Both represent the 2 shot ICL with one positive and one negative opinion presented. A plus sign indicates positive opinion, a minus sign indicates negative, and a cross indicates a mixed opinion. A Y indicates the model thought this problem existed and was serious. }
\label{tab:appendix_sway}
\end{table*}

Overall, our analysis of the AAE/AAVE task reveals three major observations. The quantifying numbers are included in Table \ref{tab:dialect_summary}, and the opinion sways are shown in Table \ref{tab:appendix_sway}.

\begin{itemize}
    \item ICL can significantly improve the amount of usage of AAE/AAVE \textit{(p < 0.05)}
        
    \item Opinions of LLMs can be swayed with ICL task, but biases cannot. 

    \item The internal rhetoric of models while using vernacular often resorts to stereotypes.  
    
\end{itemize}

\subsubsection{Prompt-only}

With just the instruction in the prompt, models establish baseline opinions for each topic, as discussed below. Throughout, we observe a recurring theme of using stereotyped African American Vernacular English (AAVE). Instead of words like \textit{alright, sure, them, they, nothing}, models resort to \textit{aight, fo sho, dey, dem, nothin'}.

    \paragraph{Llama-3:} Similar to the GPT models, Llama-3 exhibited comparable trends in opinion but demonstrated slightly negative views toward law enforcement.

    \paragraph{GPT-3.5, GPT-4, GPT-4o:} These GPT models provided structured responses incorporating informal and AAVE expressions. They attempted to discuss both positive and negative aspects of most topics, except racism and the Black Lives Matter (BLM) movement.

    \paragraph{Mistral Instruct 7B:} This model responded to most topics similarly to the GPT models. Additionally, it generated answers with implicit stereotypes of African Americans (e.g., supportive views on marijuana use within African American communities).

\subsubsection{1-shot}
In the one-shot setup, we observed a sway based on the speaker's positive or negative opinion; however, the implicit stereotype of African Americans remained. All models remained neutral on marijuana when prompted with negative opinions. Attitudes toward racism remained consistent across all models and setups, regardless of the text provided.

    \paragraph{Llama-3:} Llama-3 reflected opinions from the texts on all topics except racism and marijuana. It consistently addressed systemic racism in responses, regardless of the text provided.

    \paragraph{GPT-3.5, GPT-4, GPT-4o:} These models mirrored positive opinions from the provided texts for all topics, while negative opinions were expressed on law enforcement and BLM topics. They remained neutral on affirmative action when prompted with negative opinions. When discussing marijuana, the GPT models emphasized its impact on the Black community.

    \paragraph{Mistral Instruct 7B:} This model behaved similarly to Llama-3.

\subsubsection{2-shot}
In the two-shot responses, we consistently noted positive opinions on marijuana but unchanged responses on racism across setups. GPT-4 featured more fictional anecdotes to support marijuana use. AAVE expressions were more prevalent in two-shot responses across most models.

    \paragraph{Llama-3:} Llama-3 reacted neutrally to the topics of law enforcement and BLM. It generated slightly negative opinions on the affirmative action topic but positive opinions on marijuana. In this setup, Llama-3 offered a more in-depth discussion about systemic stereotyping, using more formal and standard English, although it remained highly repetitive.

    \paragraph{GPT-3.5, GPT-4, GPT-4o:} The GPT models were neutral on law enforcement and affirmative action but showed support for marijuana (e.g., GPT-4: "Just another gift from Mother Earth") and BLM. GPT-4 and GPT-4o adopted a conversational tone with personal anecdotes, including fabricated characters (e.g., GPT-4o: "My cousin, for example, he's a cop and he's doing his best to help the community"). GPT-4 featured more AAVE than other models.

    \paragraph{Mistral Instruct 7B:} This model exhibited a negative attitude towards law enforcement ("defunding and abolishing police") and a positive attitude towards marijuana, while remaining neutral towards affirmative action and BLM.

\section{Prior Work}
\label{sec:lit_review}

Very recently, \citet{liu-etal-2024-step-step} showed that specific region editing of generated text is a more controllable method to transfer the styles of seven tasks ranging from sentiment to formality. However, they do not show the efficacy of their model for controlling reading level or dialect. Style control has also been achieved by using GANs \citep{aich-etal-2022-demystifying}, or LLMs \citep{yang2018unsupervised}, or separately by using schema-guidance \citep{tsai2021style}. However, the effectiveness of prompts or ICL for style control has not been investigated at length. 

LLMs have also recently been used for creating teaching applications and classroom guidance \citep{xiao-etal-2023-evaluating}, as a teaching assistant \citep{hicke2023aita}, or for direct tutoring \citep{liang2023let}. While \citet{liang2023let} introduced the tailoring of exercises to a student's need, no recent teaching application of LLMs have focused on the stylistic need of the user, be it through grade-level or culturally-relevant language. 

Cultural alignment for LLMs has also been a recent area of study, with \citet{lin2023taiwan} creating a model that shows better generative capabilities with the cultural context of the end-users in mind. However, the limitations of LLMs in cultural sensitivity have been noted very recently \citep{yao2024benchmarking}. We show in this paper that while ICL and prompting can lexically improve the use of vernacular, they actually result in a distorted representation of culture through dialect. 

\section{Conclusion}
\label{sec:conclusion}

This paper aimed to demonstrate how focused prompts and properly referenced in-context learning (ICL) paradigms can control LLM-generated text for reading level and dialect. Comparing Table \ref{tab:reading_level_all} Table \ref{tab:rl_baseline}, and the baselines in Table \ref{tab:dialect_summary} reveals significant improvements in both tasks. The mean reading level decreases from 12.7 to 3.2 after prompting, and to 5.2 after one-shot and two-shot ICL\footnote{These numbers exclude Llama-3 due to high inconsistencies across all modes for that model}. Similarly, the use of AAE increases from a mean of 0.02 to 0.26 with prompting, to 0.2 with one-shot ICL, and to 0.22 with two-shot ICL, a mean increase of 0.21 points (\textit{p = 0.007}).

However, there are clear limitations in style control using prompts. \citet{brown2020language} suggested that large language models can learn tasks from few examples. The question then arises: why does performance degrade (albeit insignificantly) during ICL compared to prompting? \citet{qin2023toolllm} suggest that instruction following is complex, as user instructions are often more complicated than those seen during training. We observe similar patterns, particularly in ICL tasks.

During both one-shot and two-shot ICL, models tend to use comparisons and direct references from the provided stories, inadvertently increasing the reading level. In the AAE/AAVE one-shot task, we notice opinion sways based on the positive or negative nature of the reference. These changes however, are not significant. Furthermore, some internal stereotypes and biases persist regardless of ICL. Despite using reference texts from normal speech in interviews, all models employ more stereotypical language, including when ICL is used.

\paragraph{The main conclusions of the paper are as follows:}

\begin{itemize}
\item Prompting and In-context learning can control for both reading level and dialect - and significantly improve LLM performance. Tables \ref{tab:reading_level_all}, \ref{tab:dialect_summary} , \ref{tab:rl_baseline}
\item Opinions sway, biases don't - Language models often change perspectives and opinions on matters based on the ICL reference. However, they do not correct biases or stereotypes. Table \ref{tab:appendix_sway} and section \ref{sec:results}
\item Instruction tuning is more effective than model parameter size increases for style control tasks. Tables \ref{tab:reading_level_all} and \ref{tab:dialect_summary}
\item Additional de-biasing methods, better instruction tuning with complicated instructions, and bias checks at inference time are all essential nowadays.
\end{itemize}

Therefore in conclusion, this paper shows the challenges encountered when controlling style and dialect for large language models. Controlling for style and dialect have numerous purposes. These range from the potential to improve academic outcomes to being used in a variety of fields such as IRB forms, healthcare, medical question answering and so on. However, as we showed in this paper, there is a lot to fix before we can trust LLMs for style control. These include fixing inconsistencies, better instruction tuning, and additional guard rails at inference time. We notice that while some problems, like inconsistent responses, are local to a model (like llama). Other problems, like racist language use, is common across all model families. This highlights the need for proper tuning across all models and architectures. We hope this paper will serve as an important lens to find areas of urgent focus for generative AI in general. 

\section{Limitations}
\label{sec:limit}

This study has several limitations. First, the concept of style is inherently broad, encompassing elements such as sentiment, formality, and clarity. However, our research focuses solely on two specific aspects: reading level and vernacular English. Specifically, within the scope of vernacular English, our study is confined to examining African-American Vernacular English (AAVE). Although there are various dialects of English, our analysis only addresses the stereotypes generated by LLMs and does not dive deep into the linguistics of dialect or vernacular.

Additionally, our study's scope is limited by its reliance on few-shot learning techniques. Future research could build on our findings by incorporating fine-tuning or meta-training of large models to explore ways to mitigate stereotypes. To Considering that most users of generative artificial intelligence are not from the computer science community and may not be familiar with advanced machine learning techniques, such as supervised fine-tuning (SFT), Model-Agnostic Meta-Learning (MAML), Meta-In-Context Learning (Meta-ICL), Reinforcement Learning with Human Feedback (RLHF),  and Retrieval-Augmented Generation (RAG), our findings highlight significant performance gaps in models currently deployed for general use.

In conclusion, while our study provides valuable insights into specific aspects of language modeling and style analysis, it underscores the necessity for further research to address the broader and more complex issues of style and bias in language models. By recognizing and addressing these limitations, future work can contribute to the development of more inclusive and accurate generative AI systems that better serve a diverse user base.

\section{Ethical Concerns}
\label{sec:ethics}

All the research reported in this paper adheres to the ACM and ACL's codes and guidelines of ethics. We do not use any human (or real participant) data. 

However, LLMs that are capable of generating biased, racist, and stereotyped speech and are being used by the general population is a cause for concern. There can be many downstream detrimental effects. A reinforcement of stereotypes and discrimination can occur if group-specific stereotypes are propagated among people. A general erosion of trust in AI technologies, in an already polarized landscape. There also exists a potential for negative impact on marginalized communities. Therefore, there needs to be additional guard rails which are implemented and maintained. 

IBM recently published an online blog demonstrating how artificial intelligence (AI) bias can have significant real-world impacts across various domains, including online advertising and healthcare systems\footnote{\url{https://www.ibm.com/blog/shedding-light-on-ai-bias-with-real-world-examples/}}. Ensuring that large language models (LLMs) and generative AI systems accurately reflect the diverse variations and nuances of the real world is critical for achieving more equitable outcomes in an increasingly AI-driven society. To address this issue, two essential measures must be taken: improving the training processes of modern AI models and conducting comprehensive evaluations of their performance in real-world tasks.

It is crucial to handle the infusion of human style qualities, such as reading level and vernacular English, with the utmost sensitivity. Biases in training data, misclassifications in downstream tasks, and reliance on outdated social constructs (e.g., binary gender) are just a few ways automated systems can fail and further marginalize vulnerable populations~\cite{sap-etal-2019-risk,gonen2019lipstick}. The two models used in this study may be trained on language from non-representative samples and, thus, may fail to generalize across other populations. However, we reemphasize, that there are benefits to style control as mentioned above. Furthermore, without imparting social and cultural norms into NLP systems, we may run the risk of limited utility in NLP systems~\cite{hovy2015tagging}.

Finally, it is important to avoid anthropomorphizing dialog systems, as this can lead to transparency and trust issues, particularly in high-stakes settings (see ~\citet{abercrombie2023mirages} for an in-depth discussion). 

\section*{Acknowledgements}

This research was supported in part by the Intramural Research Program of the NIH, National Institute on Drug Abuse (NIDA).

\bibliography{custom}

\appendix
\section*{Appendix A: Baseline Results for Tasks}


\begin{table*}[h]
    \centering
    \begin{tabular}{c|c|c|c|c|c|c|c}
    \toprule
       Model &  Sun & Human Body & Plant & Water Cycle & Pollution & Mean & Std Dev\\
       \midrule
        GPT-4 & 14.50& 8.35& 9.02& 11.26& 13.80& 11.38&2.46\\
        GPT -4o & 15.25& 18.1& 13.94& 12.87&  24.95& 17.02&4.33\\
        GPT -3.5 & 10.89& 11.25& 8.62& 11.65&  17.06& 11.89&2.78\\
        LLama-3 & 4.5& 6.46& 32.45& 11.17& 48.73& 20.6& 17.2\\
        Mistral-7 & 8.92& 8.92& 7.17& 10.69& 17.16& 10.57& 3.47\\
         \bottomrule
    \end{tabular}
    \caption{Baseline Results - for reading level. These results are from when LLMs are only asked the question and nothing else}
    \label{tab:rl_baseline}
\end{table*}

\end{document}